\documentclass[runningheads]{llncs}
\usepackage{graphicx}
\usepackage{amsfonts}
\usepackage{my_ref}
\usepackage{adjustbox}
\usepackage{subcaption}

\begin{document}
\title{A Contrastive Distillation Approach for Incremental Semantic Segmentation in Aerial Images}
\titlerunning{Contrastive Distillation for ICL Semantic Segmentation in Aerial Images}

\author{Edoardo Arnaudo\inst{1,2}\orcidID{0000-0001-9972-599X} \and
Fabio Cermelli\inst{1}\orcidID{0000-0001-7077-697X} \and
Antonio Tavera \inst{1}\orcidID{0000-0002-9013-4007} \and
Claudio Rossi \inst{2}\orcidID{0000-0001-5038-3597} \and
Barbara Caputo\inst{1}\orcidID{0000-0001-7169-0158}}

\authorrunning{Arnaudo et al.}

\institute{Politecnico di Torino, Italy
\email{\{name.surname\}@polito.it}\\
\and
LINKS Foundation, Torino, Italy
\email{\{name.surname\}@linksfoundation.com}}

\maketitle

\begin{abstract}
Incremental learning represents a crucial task in aerial image processing, especially given the limited availability of large-scale annotated datasets.
A major issue concerning current deep neural architectures is known as catastrophic forgetting, namely the inability to faithfully maintain past knowledge once a new set of data is provided for retraining.
Over the years, several techniques have been proposed to mitigate this problem for image classification and object detection. However, only recently the focus has shifted towards more complex downstream tasks such as instance or semantic segmentation.
Starting from incremental-class learning for semantic segmentation tasks, our goal is to adapt this strategy to the aerial domain, exploiting a peculiar feature that differentiates it from natural images, namely the orientation. In addition to the standard knowledge distillation approach, we propose a contrastive regularization, where any given input is compared with its augmented version (i.e. flipping and rotations) in order to minimize the difference between the segmentation features produced by both inputs. We show the effectiveness of our solution on the Potsdam dataset, outperforming the incremental baseline in every test.\footnote{Code available at: \url{https://github.com/edornd/contrastive-distillation}}

\end{abstract}

\section{Introduction}
Semantic Segmentation represents a key task in aerial image processing, given its wide range of applications, from urban contexts \cite{ref_isprs}, land cover and monitoring \cite{ref_multiscale_fusion_aerial} or agricultural settings \cite{ref_agrivis_switchnorm}.

However, the majority of state-of-art solutions are designed to perform on a static set of categories by means of a full end-to-end training, with no option to integrate new knowledge. Without precautions, deep neural networks tend in fact to forget previously acquired information when a new training set is provided, resulting in poor performance on the old classes.

This phenomenon, known as \textit{catastrophic forgetting} \cite{ref_catastrophic_forg}, has been addressed and successfully mitigated through a range of different methods \cite{ref_lwf,ref_icarl,ref_icl_pruning}, mostly considering image classification or object detection.
In recent years, a greater deal of effort has been put on specific downstream tasks such as semantic segmentation, with solutions involving representation consistency \cite{ref_feng_icl_aerial}, replay-based methods \cite{ref_kd_ce_aerial}, or knowledge distillation \cite{ref_mib}.
The problem of incremental learning is extremely relevant also in aerial settings where, despite the growth in resources and data, the scarcity of large-scale annotated aerial datasets remains a crucial drawback for practical applications. In fact, it is often the case that images are collected in the same geographical area \cite{ref_isprs}, or that the data itself is not immediately available, but rather acquired and processed periodically.

In this work, we propose to tackle the problem of incremental-class learning (ICL) in the context of semantic segmentation, focusing on aerial imagery. Leveraging on the MiB framework \cite{ref_mib}, a distillation-based method specifically designed for semantic segmentation tasks, we introduce an additional regularisation based on contrastive distillation, with the aim of exploiting a distinctive feature of such images, namely their invariance to orientation.
We explicitly model this feature by comparing the activations produced by the framework on the input and its transformed version, minimising their difference. A first step involves the student network, comparing pairs of augmented inputs, then activations are also compared with the teacher from the previous incremental step, to improve the knowledge distillation.
We evaluate our solution on the Potsdam benchmark dataset \cite{ref_isprs}, where it consistently outperforms the robust incremental baseline in every setting.
In summary, our contributions can be listed as follows:
\begin{itemize}
    \item We address the problem of ICL in semantic segmentation of aerial images, providing benchmark results on a popular dataset.
    \item We propose a new regularization and distillation approach based on contrastive representation learning, addressing the arbitrary orientation of the inputs, one of the key aspects of aerial images.
\end{itemize}

\vspace{-10pt}
\section{Related Work} \label{sec:related}
\vspace{-5pt}

\subsubsection{Aerial Semantic Segmentation.} \label{sssec:related_aerial_seg}

Thanks to the recent advancements in deep learning, many semantic segmentation approaches have been proposed over the years \cite{ref_fcn,ref_deeplab,ref_pspnet,ref_resunet_a}, focusing mostly on natural images. Most common methods revolve around encoder-decoder, fully-convolutional architectures \cite{ref_deeplabv3plus}.
These techniques have been successfully applied to the field of aerial images in wide range of contexts, such as semantic labelling in urban \cite{ref_beyond_rgb,ref_resunet_a,ref_seg_aerial_nogueira} or agricultural scenarios \cite{ref_agrivis_switchnorm,ref_seg_aerial_nogueira}, or land cover tasks \cite{ref_isprs_late_fusion}. Despite the strong similarities with the natural counterparts, aerial images present some peculiar differences that have been addressed with varying approaches: first, satellite imagery are seldom limited to the visible spectrum and often include additional frequencies \cite{ref_agrivis_switchnorm}. Common solutions to this problem include simpler solutions such as the duplication of input weights \cite{ref_seg_coinnet} or finer multi-modal approaches comprising the fusion of different modalities \cite{ref_isprs_late_fusion,ref_multiscale_fusion_aerial}.
Last, a peculiar aspect of aerial and satellite images is represented by the top-down view, in which the orientation becomes arbitrary. In our work we propose to leverage on this peculiar feature, already successfully exploited in classification tasks \cite{ref_rotinv_clf}, \cite{ref_rotinv_rim}, by applying a contrastive regularization to both the segmentation task and the incremental tasks, to further improve the knowledge distillation between steps.

\vspace{-20pt}
\subsubsection{Incremental Learning.} \label{sec:related_icl}
Catastrophic forgetting \cite{ref_catastrophic_forg}, meaning the inability to remember past knowledge upon learning new information, represents a major issue concerning current deep learning solutions. Several techniques have been proposed to mitigate this issue, with different approaches: replay-based methods \cite{ref_icarl,ref_kd_ce_aerial}, exploiting exemplars from old classes parameter isolation parameter-based methods \cite{ref_icl_pruning}, involving a selective pruning so that the weights representing old labels are maintained through the learning steps, and memory-based approaches \cite{ref_icl_synaptic}, where important parameters from previous steps are consolidated, forcing the model to maintain a robust representation for old classes.
Last, one of the most effective techniques focuses on data and exploits knowledge distillation \cite{ref_lwf,ref_mib}. The latter is usually carried out with a \textit{teacher-student} approach.
Considering Semantic Segmentation on aerial imagery, a first proposal is represented by \cite{ref_kd_ce_aerial}: here, an hybrid approach comprising both knowledge distillation and additional supporting exemplars is employed. Similarly, in \cite{ref_feng_icl_aerial} the distillation approach is improved by strengthening the internal representations throughout the learning steps.
Compared to image classification, semantic segmentation presents peculiarities that may lead to poor performances when not addressed, such as the presence of a common \textit{background} label.
In MiB \cite{ref_mib}, this issue is tackled by taking into consideration this distributional shift, by means of unbiased losses and regularizations with respect to the background label.

\vspace{-10pt}
\subsubsection{Contrastive Learning.} \label{sec:related_crl}
Contrastive learning has become one of the most promising recent techniques in deep learning, closing the gap between supervised and self-supervised settings \cite{ref_ssl_pirl,ref_ssl_simclr,ref_ssl_moco,ref_ssl_swav}, or even improving the former by learning more robust representations \cite{ref_sup_contrastive_neurips2020}.
The objective of Contrastive Representation Learning (CRL) is to cluster together latent representations of similar samples (i.e. \textit{positive examples}), while at the same time increasing the distances between instance representations of different categories \textit{(i.e. negative examples)}. 
CRL is often applied exploiting pretext tasks (i.e. manually devised tasks solely based on the image itself), including: geometric or color transformations \cite{ref_ssl_pirl,ref_ssl_simclr}, image reconstruction from its parts \cite{ref_ssl_jigsaw,ref_ssl_gan_segmentation}, or cross-modal techniques \cite{ref_ssl_intermodal_rot,ref_ssl_multimodal_semseg}.
These additional tasks can also be paired with more traditional supervised settings such as semantic segmentation, in order to improve the results on the main task \cite{ref_ssl_gan_segmentation,ref_ssl_multimodal_semseg}, deal with low resource datasets \cite{ref_contrastive_seg_limited}, or integrate additional modalities \cite{ref_comir_nips2020,ref_ssl_multimodal_semseg}.
Here, we propose a similar approach where the same inputs are augmented twice, however we exploit the resulting representations as a further regularization to induce further invariance with respect to the applied transformations, during both standard training and knowledge distillation.

\section{Methodology}

\subsection{Problem statement}
\vspace{-5pt}
We address the problem of Incremental-Class Learning (ICL) for Semantic Segmentation on aerial images, where we suppose that different portions of data are provided sequentially, each one with a different set of labels.\\
First, we can define Semantic Segmentation as a pixel-wise classification, where each pixel $x_i$ composing a generic image $x \in X$ with constant dimensions $H \times W$, is associated with a label $y_i \in Y$ representing its category, or eventually associated with a generic and comprehensive \textit{background} class $b \in Y$.
The training can be defined as learning a model $f_\theta$ with parameters $\theta$, mapping from the image space $X$ to the pixel-wise label space $Y$, namely: $f_\theta: X \mapsto \bbbr^{|H \times W \times Y|}$.

Considering now the ICL setting, we require multiple sequential training phases named \textit{learning steps}, in which we provide a different set of data samples and labels every time.
Specifically, at each step $t$, we expand the previous set of labels $Y^{t-1}$ with the additional ground truth $Y^t$, obtaining a new set of labels $C^t = Y^{t-1} \cup Y^t$. At each phase, we are also provided with a new training set $D^t$, such that each pixel-wise label $y_i$ belongs to one of the current categories $Y^t$ or the generic background class $b$.
We then train a new model $f_\theta^t$ on the whole set of categories $C^t$, deriving the old labels from the outputs of the previous model $f_\theta^{t-1}: X \mapsto \bbbr^{|H \times W \times Y^{t-1}|}$ and the new labels via standard training, exploiting the dataset for the current step.
The final goal is to obtain a single model, able to perform well on both and new classes, namely $f_\theta^{t}: X \mapsto \bbbr^{|H \times W \times C^{t}|}$.

\vspace{-10pt}
\subsection{Baseline}
\vspace{-5pt}
As previously mentioned, we adopt MiB as robust incremental baseline \cite{ref_mib}.
In ICL applied in the context of image classification, a standard approach involves a two-way training, combining a supervised loss on the dataset at the current step $D_t$ with an additional term to maintain the old knowledge. In the case of the selected framework, the latter is carried out through distillation of the old model's outputs.
Specifically, the final loss at each learning step becomes:
\begin{equation}
    \vspace{-5pt}
    L(\theta^t) = L_{CE}(\theta^t) + \lambda L_{KD}(\theta^t)
\end{equation}

Where $L_{CE}(\theta^t)$ represents a supervised Cross-Entropy loss, while $L_{KD}(\theta^t)$ represents the Knowledge Distillation term at step $t$ from the previous model $f_{\theta^{t-1}}$, weighted by a factor $\lambda$.

As briefly stated in \cref{sec:related_icl}, it is common that two sets of categories, namely $Y_i$ and $Y_j$ share the common background class $b$, however the semantic regions of the image are assigned to such label is often different in every set.
This aspect of semantic segmentation needs to be dealt with during the incremental steps, taking into account that a pixel labeled as background in the dataset $D_t$ might instead belong to one of the previous classes from step $0$ to $t-1$. Thus, for each pixel $i$ of a generic image $x$, the predicted probability $q(i, b)$ for the background class is substituted with:

\begin{equation}
    \vspace{-5pt}
    q(i, b) = \sum_{k \in Y^{t-1}} q_x^t(i, k)
\end{equation}

In other words, the background is not considered as a category on its own, but rather a probability of having an old class \textit{or} actual background.

A similar concept is adopted for the distillation component, where the following distillation loss is applied:

\begin{equation}
    \vspace{-5pt}
    \label{eq:kdloss}
    L_{KD}^{\theta^t}(x, y) = \frac{1}{N} \sum_{i \in x} \sum_{c \in Y^{t-1}} q_x^{t-1}(i, c) log(q_x^t(i, c))
\end{equation}

Where the last term refers to the predicted probabilities for the new model with respect to the old classes. Given that the contribution for the new labels is provided by the Cross-Entropy loss, we require that $q_x^t(i, c) = 0$, $\forall c \in Y^t \setminus \{b\}$.
In every other case, the term represents the predicted probability for the new model of having a label $c$ for a pixel $i$, normalized across old classes as reported in \cite{ref_mib}.
Again, the distributional shift of the background class needs to be addressed for the incremental learning as well. Consequently, the predicted probability $q_x^t(i, b)$ for this special class is rewritten as:

\begin{equation}
    \vspace{-5pt}
    q_x^t(i, b) = \sum_{k \in Y^t} q_x^t(i, k)
\end{equation}

In other terms, the predicted probability for the background class of the new model is substituted with the probability of having a new class \textit{or} the background. In fact, we expect regions belonging to the new classes to be ignored by the previous model, thus labelled as generic background.

Moreover, excluding similarities among categories, it is extremely likely that predictions for $f_{\theta^{t-1}}$ for the current classes $Y^t$ will fall under the background class. For this reason, we perform the same weight initialization for the final classifier as proposed in \cite{ref_mib}, so that its outputs for the new classes are uniformly distributed around the background from the very beginning to ease convergence.

\vspace{-10pt}
\subsection{Contrastive Distillation}
\vspace{-5pt}
\begin{figure}[ht]
\includegraphics[width=\textwidth]{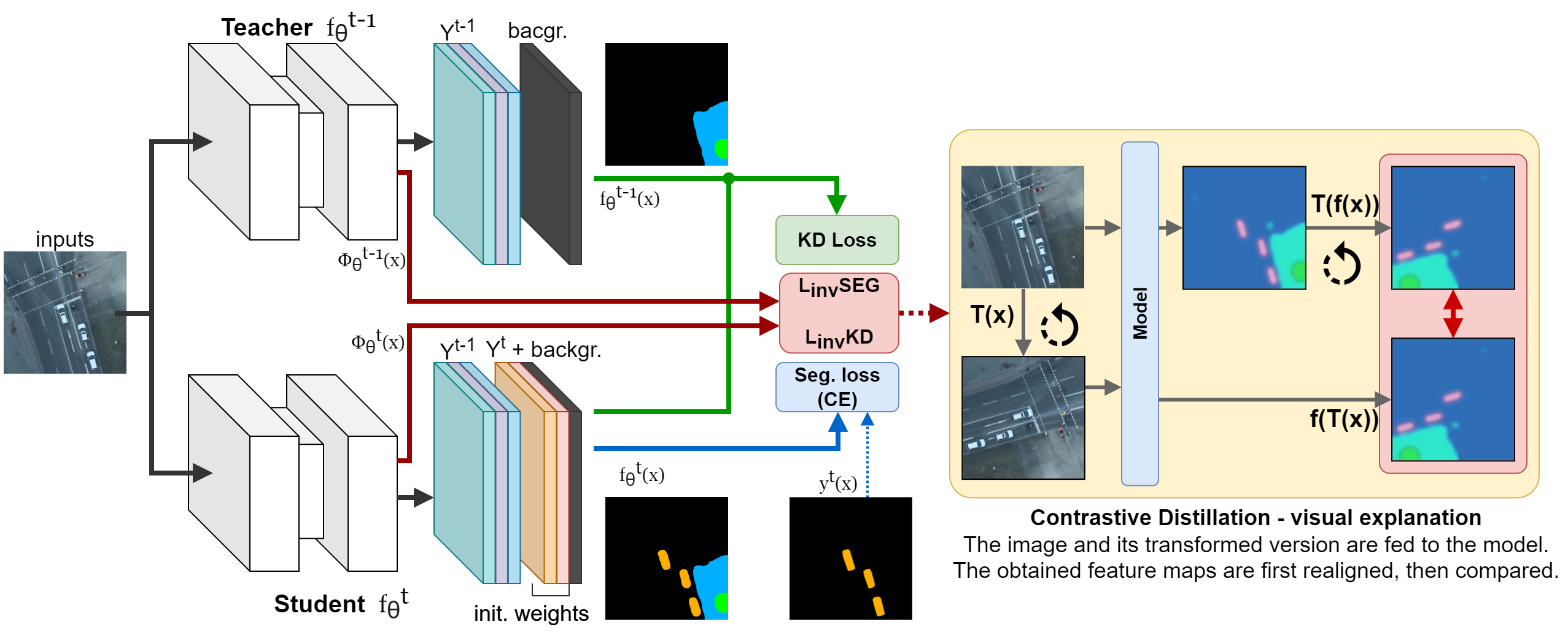}
\caption{\small{Overview of the ICL setting on aerial images. For each step $t$, both the image $x$ and its augmented version $T(x)$ are provided to the old (top) and new (bottom) models. New classes are trained with supervised training on the available ground truth (blue), while old categories are learned through KD (green). Last, features of the augmented inputs are confronted with the augmented features of the normal input, on both distillation and supervised training (red).}} \label{fig:schema}
    \vspace{-15pt}
\end{figure}

As stated in \cref{sssec:related_aerial_seg}, a major difference between natural and aerial images is represented by their orientation: in the former case, the point of view is fundamental to the correct detection of an entity. In fact, in a common scenario we expect to find background and foreground entities in a specific part of the image (e.g. animals in a specific pose, sky on top, ground on the bottom). In the latter case instead, given that the orientation is often arbitrary and simply given by the direction of the observation mean, image rotations around the top-down axis become meaningless for the correct classification or detection.

Therefore, we explicitly model this orientation bias by introducing an additional regularization, both to the supervised training and the incremental knowledge distillation, using a contrastive-based approach.
Specifically, given a generic input image $x$, we can obtain the output features of the current model $\phi_{\theta}(x)$ (thus excluding the final classifier). At the same time, given the same image transformed with augmentation $T$, the model should output a new activation, namely $\phi_\theta(T(x))$. Given the invariance to rotation, we can assume that both outputs are comparable, minus a transformation, which can be directly applied to the first activation.
Formally, we can therefore introduce a regularization term at each learning step $t$, namely $L_{inv}^{SEG}$ as:

\begin{equation}
    L_{inv}^{SEG} = MSE(\phi_{\theta^t}(T(x)), \ T(\phi_{\theta^t}(x))
    \label{eq:auginv-ce}
\end{equation}

In other words, the additional term minimizes the differences between \textit{the features of the model on the transformed image and the transformed features of the same model on the original image}, exploiting a Mean Squared Error between the features.

In an ICL setting, we are also interested in transferring the knowledge between $f_{\theta^{t-1}}$ and $f_{\theta^{t}}$, so that the previous outputs are maintained as unaltered as possible. Together with the standard KD loss from \cref{eq:kdloss}, we can apply the same invariance principle between old and new models. More formally, at each step $t > 0$ we can introduce a further regularization as:

\begin{equation}
    L_{inv}^{KD} = MSE(\phi_{\theta^t}(T(x)), \ T(\phi_{\theta^{t-1}}(x))
    \label{eq:auginv-icl}
\end{equation}

Simply put, this term minimizes the difference between the features of the new model derived from the transformed image and the transformed features of the old model, obtained from the non-augmented version of the input.

In summary our method comprises three regularizations, therefore the final loss to be minimized can be expressed as:

\begin{equation}
    \label{eq:final_loss}
    L(\theta^t) = L_{CE}(\theta^t) + \lambda L_{KD}(\theta^t)  + \eta L_{inv}^{SEG}(\theta^t) + \rho L_{inv}^{KD}(\theta^t), 
\end{equation}

where the terms $\lambda$, $\eta$ and $\rho$ are scalar factors, weighting the contribution of the additional losses.
The overall framework is illustrated in \cref{fig:schema}.

\vspace{-10pt}
\section{Results}

\vspace{-5pt}
\subsection{Experiments} \label{ssec:experiments}
\vspace{-5pt}

As described in the previous section, we build our method on top of the MiB framework, which represents a strong baseline for ICL in segmentation tasks.
We perform all our experiments on the Potsdam dataset \cite{ref_isprs}, a well known benchmark on aerial imagery providing an urban land cover subdivided into six classes: \textit{impervious surfaces}, \textit{building}, \textit{low vegetation}, \textit{trees}, \textit{cars} and \textit{clutter}. The dataset contains 38 large patches taken from the namesake city, where each patch has a fixed size of $6000 \times 6000$. Each patch comes with a sampling resolution of $5cm$ and provides five different modalities, namely: red (R), blue (B), green (G), infrared (IR) and a normalized digital surface map (DSM), all encoded as TIFF files.
Given our focus on ICL, we only include in our tests inputs composed of RGB and RGBIR, discarding the additional surface map.

Every incremental set of labels is assumed to be disjoint from the previous ones. However, given the aerial setting, it is quite common that each image contains many of the available labels. For this reason, we first split the set into disjoint partitions, such that each split only contains a single label. 
Formally, considering a full dataset $D \subset X \times Y^{|H \times W|}$, we subdivide the available data into $|Y|$ disjoint partitions $D_y$ such that $D_i \cap D_j = \emptyset$ $\forall i, j \in Y$ where $i \neq j$, and each partition only contains a set of labels $Y_i = \{i, b\}$, i.e the set of images is unique for each partition and each split only contributes to the whole training with a single label, or a generic background. Every incremental step will then include a variable number of classes, which will in turn require all the partitions corresponding to the involved categories.

\vspace{-10pt}
\subsection{Implementation details}
\vspace{-5pt}
For all the experiments we adopted an encoder-decoder architecture with residual connections, based on the Res-UNet model \cite{ref_resunet_a}. Since memory requirements are crucial for the incremental setting, we introduce two optimizations: first, we swap the standard ResNet backbone with an equivalent yet more efficient TResNet with ImageNet pretraining \cite{ref_tresnet}. The latter applies a series of optimizations aimed at maximizing the data throughput on GPU, while at the same time improving the performance over the classical residual architectures.
For the experiments concerning four input channels, namely RGBIR, we expand the input layers duplicating the weights of the red channels, with a similar approach to \cite{ref_seg_coinnet}.
Second, we apply in-place activated batch normalization also on the decoder, as proposed in \cite{ref_inplaceabn}, further reducing the memory footprint of the architecture. 

We train the model for 80 epochs for each step, using AdamW as optimizer with learning rate of $10^{-3}$ and a cosine annealing scheduler, while reducing to $10^{-4}$ for the last steps. We adopt a batch size of 8, with effective size equal to 16 given that the pairs generated via contrastive augmentation are also exploited for the supervised training.
Given the large size of the inputs, we tile the $6000 \times 6000$ images of the Potsdam dataset into patches with size $512 \times 512$ with overlap of 12 pixels, which is the minimum amount required to avoid partial tiles while also minimizing the replication of the image content.
We perform robust data augmentation as in \cite{ref_resunet_a} in every setting, focusing on elastic transformations.
Considering the contrastive regularization, we maintain the setting provided in \cite{ref_mib}. We set the factors $\eta = \rho = 0.1$ in every test and evaluate as transformation random vertical and horizontal flipping, with rotations by 90-degree angles.
In order to monitor the performances, we select $15\%$ of the training set as validation. The final results are reported as F1 scores on the benchmark test set.

\vspace{-10pt}
\subsection{Potsdam dataset}
\vspace{-5pt}
Given the high similarities among image patches and the uniform distribution of the labels among the tiles, the overlapped setting \cite{ref_mib,ref_kd_ce_aerial}, (i.e. images are kept even if they contain future classes), is not complex enough for a robust evaluation of the proposed regularizations.
For this reason, we implement the \textit{split} protocol described in \cref{ssec:experiments}: we first tile the original patches to obtain fixed-size input images, then we partition the dataset into 5 different disjoint sets, where each one is associated with a single label. Then, we randomly assign each tile to the smallest set among the labels present in the current tile, obtaining a uniform allocation of the data samples among the classes.
This configuration can be seen as having 5 different datasets, where each one only contains a single type of annotation. The disjoint splits ensure that the model will work on unseen images at each step, further increasing the robustness of the tests.

\begin{figure}[ht]
    \begin{subfigure}{0.5\textwidth}
        \includegraphics[width=1.0\linewidth]{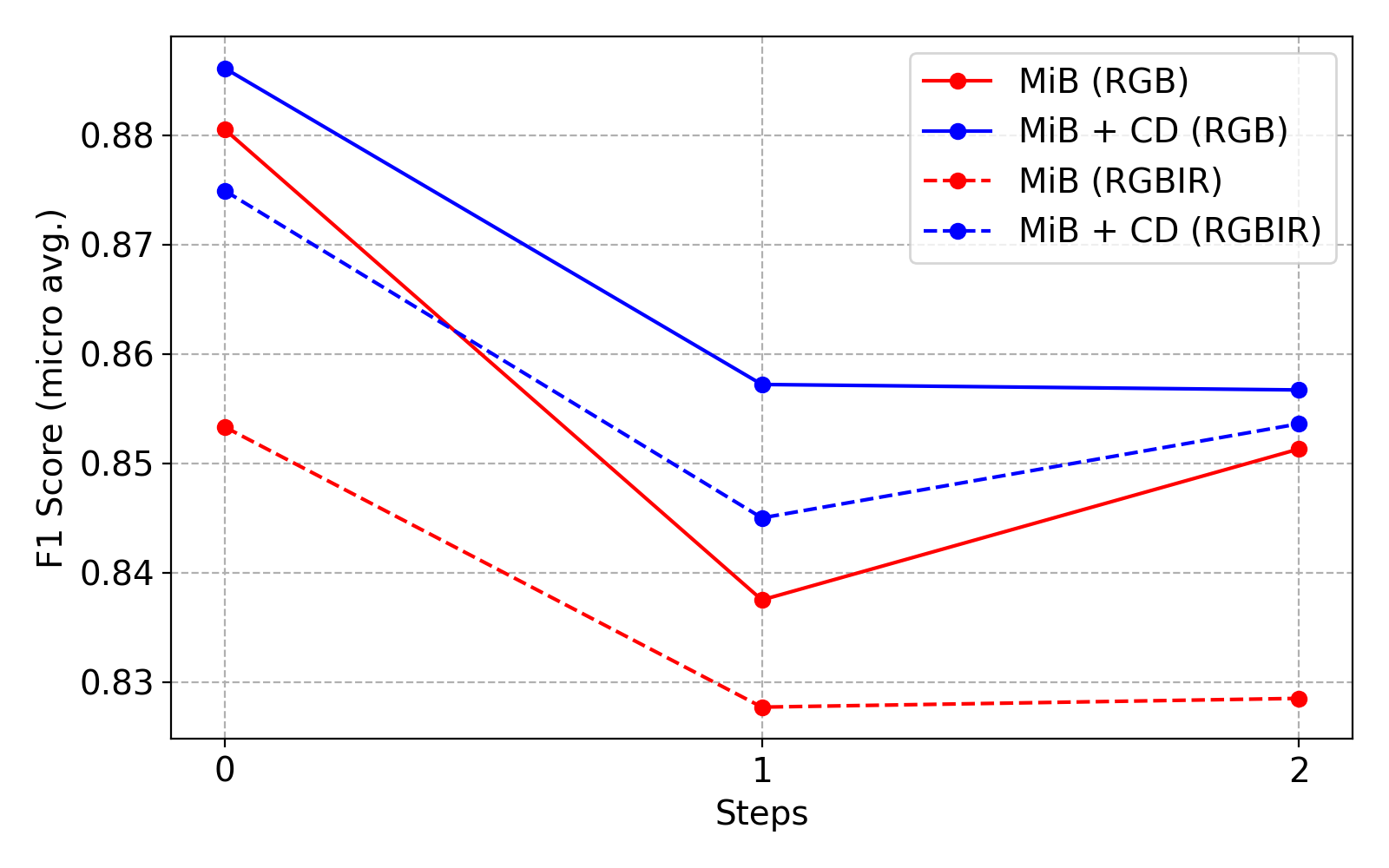} 
        \label{fig:321-results}
    \end{subfigure}
    \begin{subfigure}{0.5\textwidth}
        \includegraphics[width=1.0\linewidth]{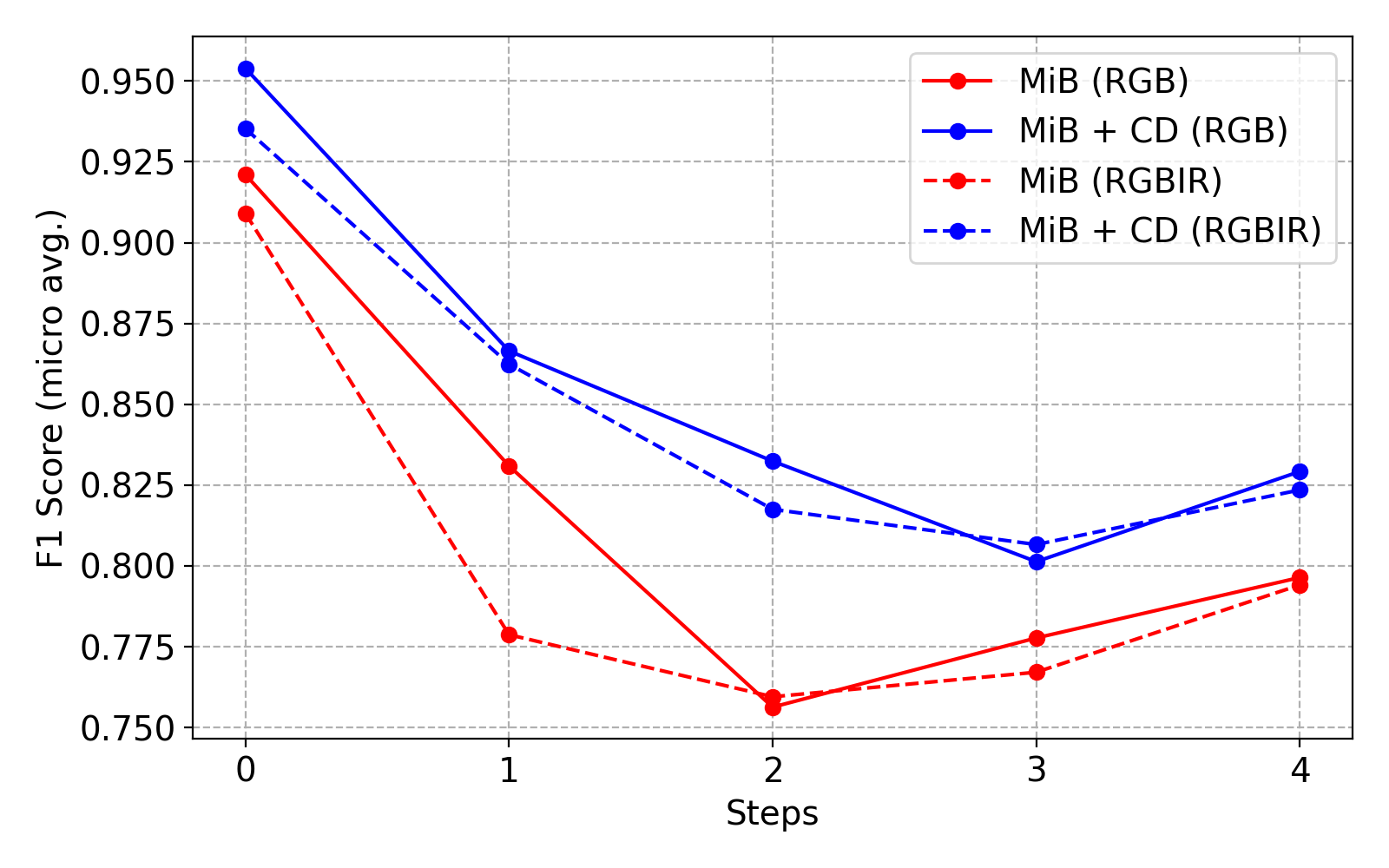}
        \label{fig:5s-results}
    \end{subfigure}
    \label{fig:icl-results}
    \vspace{-20pt}
    \caption{\small{Micro-averaged F1 scores over the incremental steps in the \texttt{3-2-1} configuration (left) and \texttt{5S} (right). Blue indicates Contrastive Distillation (CD), dashed lines the RGBIR version.}}
    \vspace{-20pt}
\end{figure}

We perform tests for two different configurations: first we replicate the testing scenario proposed in \cite{ref_kd_ce_aerial} where we suppose to receive, for the initial step, the labels for \textit{building} and \textit{trees}, then \textit{impervious surfaces} and \textit{low vegetation}, and as last step \textit{car} (\texttt{3-2-1}).
Second, we perform a more challenging test with the same order of labels, but provided sequentially (\texttt{5S}). For this last configuration, we exclude the \textit{clutter} category, since it is not included in the official benchmarks \cite{ref_isprs}.
Results for both configurations are shown in \cref{tab:results_321} and \cref{tab:results_5s}. Given the framework explicitly designed for segmentation, the MiB baseline performs reasonably well, even considering the fully sequential setting. However, the contrastive distillation approach consistently improves the performances in every experiment and every step, as reported in \cref{fig:5s-results}, even in the multi-spectral tests.
We note that in the simpler \texttt{3-2-1} setting the RGB baseline performs on par with the regularized version. We argue because of both the effectiveness of the standard approach and the robust backbone pretrained on RGB images. However, in more challenging scenarios such as \texttt{5S}, the contribution of the additional regularization is far more prominent, with a total increase over MiB of around $4\%$.

\begin{table*}[ht]
    \caption{\small{Class-wise and average results (F1 Score) obtained after 3 incremental steps (\texttt{3-2-1}). Double vertical lines indicate label groups for each step.}}
    \label{tab:results_321}
    \begin{adjustbox}{width=1.0\textwidth}
        \centering
        \begin{tabular}{|r||c|c|c||c|c||c||c|c|}
        \hline
        \multicolumn{1}{|c||}{\textbf{Method}} & \textbf{Building} & \textbf{Tree} & \textbf{Clutter}        & \textbf{Surf.} & \textbf{Low veg.} & \textbf{Car} & \textbf{Avg.} \\ \hline
        MiB (RGB)                & 0.9116            & 0.8217        & 0.2766                  & 0.8918         & 0.7589            & 0.8500       & 0.7517              \\ \hline
        MiB + CD (RGB)           & 0.9209            & 0.8085        & 0.3119                  & 0.9021         & 0.7619            & 0.8541       & \textbf{0.7599}     \\ \hline
        MiB (RGBIR)              & 0.8708            & 0.8062        & 0.2682                  & 0.8773         & 0.7414            & 0.8176       & 0.7303              \\ \hline
        MiB + CD (RGBIR)         & 0.9178            & 0.8190        & 0.3128                  & 0.8950         & 0.7635            & 0.8515       & \textbf{0.7598}     \\ \hline
        \end{tabular}
    \end{adjustbox}
    \vspace{-10pt}
\end{table*}

\begin{table*}[ht]
    \caption{\small{Class-wise and avg. results (F1 Score) obtained after 5 incremental steps (5S).}}
    \label{tab:results_5s}
    \begin{adjustbox}{width=1.0\textwidth}
        \centering
        \begin{tabular}{|r||c|c|c|c|c||c|c|}
            \hline
            \multicolumn{1}{|c|}{\textbf{Method}} & \textbf{Building} & \textbf{Tree}   & \textbf{Surfaces} & \textbf{ Low veg.} & \textbf{Car} & \textbf{Avg.} \\ 
            \hline
            MiB (RGB)                    & 0.8451   & 0.7449 & 0.7912        & 0.7011   & 0.6759 & 0.7810            \\ \hline
            MiB + CD (RGB)               & 0.9015   & 0.7515 & 0.8848        & 0.7313   & 0.8287 & \textbf{0.8195}   \\ \hline
            MiB (RGBIR)                  & 0.8564   & 0.7007 & 0.8575        & 0.6862   & 0.8228 & 0.7847            \\ \hline
            MiB + CD (RGBIR)             & 0.8770   & 0.7740  & 0.8755       & 0.7343   & 0.8437 & \textbf{0.8209}   \\ \hline
        \end{tabular}
    \end{adjustbox}
\end{table*}

\subsection{Ablation study}
In \cref{tab:ablation}, we report an ablation study highlighting the contribution of our proposals, on the \textit{split 5S} configuration with RGB input. We first start from a simple finetuning (FT): as expected, a new training without considering previous knowledge is detrimental for every step but the last.
We then test on the MiB framework that already provides excellent results, with an average increment of more than $60\%$ over the simple FT baseline.
Naively introducing the single $L_{inv}^{SEG}$, (i.e. acting on the current step only) results in better scores for the last class, as expected. However, this negatively affects the performance on previously seen categories, which are not taken into consideration.
On the other hand, applying the single $L_{inv}^{KD}$ between current and old model allows for higher scores for previous categories, increasing the average score of $2\%$, though without obtaining any boost on the labels for the current step.
Combining the two regularizations, it is possible to both improve over the current step and increase the performance over old classes, with a significant boost of around $4\%$ over the strong MiB baseline and close to the theoretical upper bound of the \textit{offline} test, representing a static multi-class learning over the whole set at once.
As additional test for the distillation capabilities of the regularization, in the second row of \cref{tab:ablation} we report results for finetuning, using both the unbiased cross-entropy from \cite{ref_mib} and CD, without actual distillation loss. The scores confirm that the additional losses actively contribute in maintaining previous knowledge.

\begin{table*}[ht]
    \vspace{-20pt}
    \caption{\small{Ablation study applied to the $5S$ setting, as class-wise F1 Scores of the last incremental step and averaged across classes.}}
    \label{tab:ablation}
    \begin{adjustbox}{width=1.0\textwidth}
    \centering
    \begin{tabular}{|l||c|c|c|c|c||c|}
    \hline
    \textbf{Method}                          & \textbf{Building} & \textbf{Tree} & \textbf{Imp. surf.} & \textbf{Low veg.} & \textbf{Car} & \textbf{Average} \\ \hline
    FT                                                                         & 0.0                 & 0.0             & 0.0                   & 0.0                 & 0.8708       & 0.1742           \\ \hline
    FT, Unb. CE + CD                                                           & 0.6118	           & 0.4927        & 0.6924	             & 0.2909            & 0.5275       & 0.5231      \\ \hline
    MiB                                                                        & 0.8491            & 0.7625        & 0.8480              & 0.6751            & 0.7703       & 0.7810           \\ \hline
    MiB + $L_{inv}^{SEG}$                                & 0.8178            & 0.7452        & 0.8514              & 0.6781            & 0.8186       & 0.7822           \\ \hline
    MiB + $L_{inv}^{KD}$                                   & 0.9079            & 0.7522        & 0.8815              & 0.7011            & 0.7895       & 0.8064           \\ \hline
    MiB + $L_{inv}^{SEG} + L_{inv}^{KD}$ & 0.9015            & 0.7515        & 0.8848              & 0.7313            & 0.8287       & \textbf{0.8196}           \\ \hline
    \hline
    Offline & 0.9510 & 0.8535 & 0.9063 & 0.8415 & 0.8942 & 0.8893 \\ \hline
    \end{tabular}
    \end{adjustbox}
\end{table*}

\begin{figure}[!ht]
    \centering
    \includegraphics[width=0.95\linewidth]{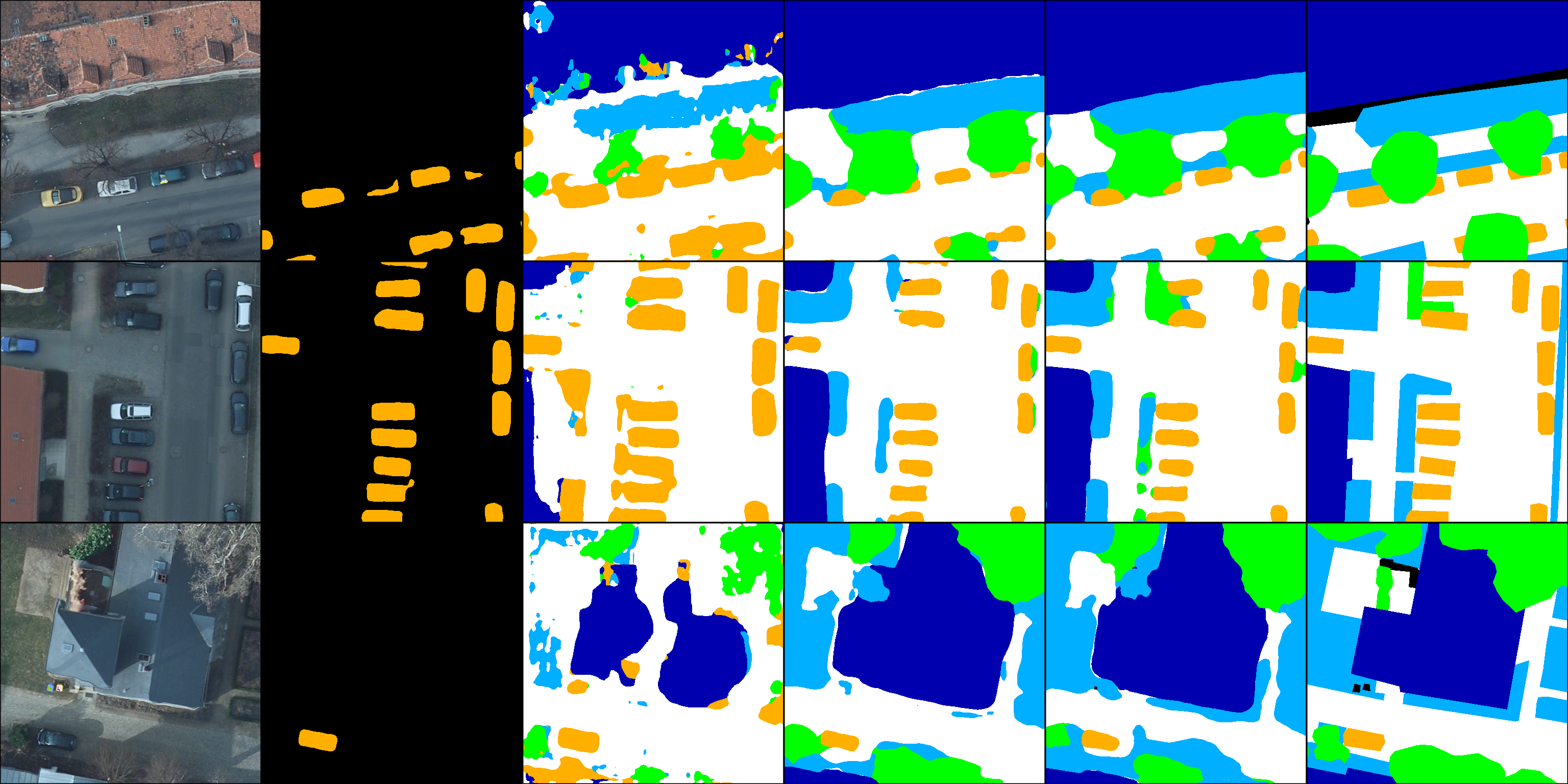} 
    \label{fig:examples}
    \caption{\small{From left to right: input, finetuning (FT), Finetuning with unbiased CE and Contrastive Distillation (FT + CD), Modelling the Background (MiB), MiB with Contrastive Distillation (MiB + CD), ground truth.}}
    \vspace{-10pt}
\end{figure}

\section{Conclusions}
\vspace{-10pt}
We addressed the problem of incremental learning in the context of semantic segmentation of aerial imagery, proposing a new regularization based on contrastive distillation to explicitly model the orientation invariance of such top-down images.
In our experiments, we first provide benchmark results for the current state-of-the-art technique on natural images, already displaying excellent performances. We then demonstrate the effectiveness of our simple additional solution leveraging on the same framework, that consistently outperforms the strong baseline leading to a more stable sequential training.
Nevertheless, incremental learning remains a challenging problem, especially considering different data sources and domains. Future works could provide more insight on this technique with additional datasets and explore more diverse scenarios, where datasets not only come with different annotations, but also from different domains.

\subsubsection{Acknowledgements}
This work was developed in the context of the Horizon 2020 projects SHELTER (grant agreement n.821282) and SAFERS (grant agreement n.869353).

\vspace{-5pt}

\bibliographystyle{splncs04}
\bibliography{refs}

\end{document}